\documentclass[sigconf,screen]{acmart}
\setcopyright{none}
\copyrightyear{2026}
\acmYear{2026}
\acmConference[ACM AI Summit '26]{ACM AI Leadership Summit}{2026}{}
\acmBooktitle{ACM AI Leadership Summit 2026}
\acmDOI{}
\acmISBN{}
\usepackage{amsmath}
\usepackage{algorithm}
\usepackage{algpseudocode}
\usepackage{tikz}

\definecolor{FigBlue}{HTML}{4285F4}
\definecolor{FigRed}{HTML}{EA4335}
\definecolor{FigYellow}{HTML}{FBBC05}
\definecolor{FigGreen}{HTML}{34A853}

\definecolor{googleblue}{HTML}{4285F4}
\definecolor{googlered}{HTML}{EA4335}
\definecolor{googleyellow}{HTML}{FBBC05}
\definecolor{googlegreen}{HTML}{34A853}

\usepackage{booktabs}
\usepackage[table]{xcolor}

\usepackage{tikz}
\usetikzlibrary{arrows.meta, backgrounds, fit, positioning, calc}
\usepackage{xcolor}

\DeclareMathOperator{\TopOp}{Top}
\DeclareMathOperator{\Nov}{Nov}
\DeclareMathOperator{\Rel}{Rel}
\DeclareMathOperator{\Feas}{Feas}
\DeclareMathOperator{\Execute}{Execute}
\AtBeginDocument{%
  \providecommand\BibTeX{{%
    \normalfont B\kern-0.5em{\scshape i\kern-0.25em b}\kern-0.8em\TeX}}}
\begin{document}

\title{Personalized Auto-Research: Towards a True AI Co-Scientist}

\author{Bo Ni}
\email{bo.ni@vanderbilt.edu}
\affiliation{%
  \institution{Vanderbilt University}
  \city{Nashville}
  \state{Tennessee}
  \country{USA}}

\author{Franck Dernoncourt}
\email{dernonco@adobe.com}
\affiliation{%
  \institution{Adobe Research}
  \city{San Jose}
  \state{California}
  \country{USA}}

\author{Hongjie Chen}
\email{hojiechen@gmail.com}
\affiliation{%
  \institution{Dolby Laboratories}
  \city{San Francisco}
  \state{California}
  \country{USA}}

\author{Yu Wang}
\email{yu.wang6@uga.edu}
\affiliation{%
  \institution{University of Georgia}
  \city{Athens}
  \state{Georgia}
  \country{USA}}

\author{Nesreen K. Ahmed}
\email{n.kamel@gmail.com}
\affiliation{%
  \institution{Cisco AI Research}
  \city{San Jose}
  \state{California}
  \country{USA}}

\author{Zhengzhong Tu}
\email{tzz@tamu.edu}
\affiliation{%
  \institution{Texas A\&M University}
  \city{College Station}
  \state{Texas}
  \country{USA}}

\author{Tyler Derr}
\email{tyler.derr@vanderbilt.edu}
\affiliation{%
  \institution{Vanderbilt University}
  \city{Nashville}
  \state{Tennessee}
  \country{USA}}

\author{Ryan A. Rossi}
\email{ryarossi@gmail.com}
\affiliation{%
  \institution{Adobe Research}
  \city{San Jose}
  \state{California}
  \country{USA}}

\renewcommand{\shortauthors}{Ni et al.}

\begin{abstract}
AI co-scientists that generate hypotheses, retrieve related work, design experiments, execute code, and draft full papers are beginning to change how research is carried out. Despite this rapid progress, state-of-the-art systems remain \emph{researcher-agnostic}: given a research goal, they optimize novelty, validity, or reviewer score while ignoring the individual scientist who will use the output. This overlooks a fundamental fact about research, namely, that what counts as novel, valuable, or feasible depends on the researcher, including their prior work, methodological repertoire, and the collaborators and communities in which they are embedded. In this work, we introduce the problem of \emph{personalized auto-research}, which conditions every stage of the research process on a representation of the individual researcher. We argue that personalization is not a convenience layer, but rather the fundamental property that allows an AI system to serve as a genuine co-scientist rather than a generic instrument. To address this problem, we propose a general and flexible framework that threads a graph-grounded researcher context through retrieval, hypothesis search, experimentation, writing, and review. The framework consists of three fundamental components: (i) graph-grounded researcher representations, (ii) personalization across the full research pipeline, and (iii) evaluation grounded in the individual. Notably, we highlight a one-size-fits-all failure mode where distinct researchers issuing the same goal receive essentially the same research, erasing the tacit knowledge through which novel ideas arise. 
Finally, we discuss fundamental open problems and challenges.
\end{abstract}

\maketitle

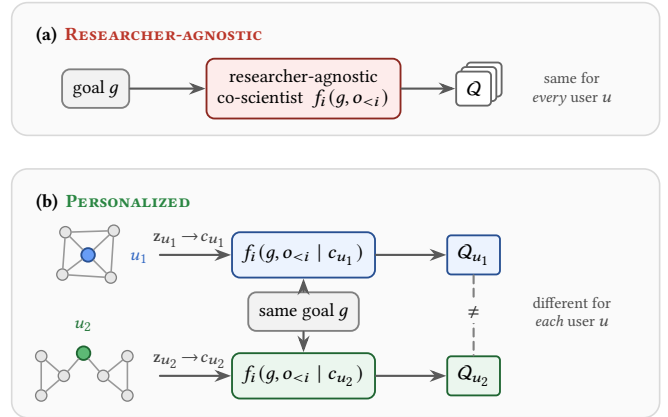
\begin{figure}[t]
\vspace{4mm}
\centering
\begin{tikzpicture}[
    font=\footnotesize, 
    line join=round, 
    line cap=round,  
    box/.style={draw=black!60, rounded corners=3pt, inner sep=4pt, align=center, line width=0.7pt},  
    goalbox/.style={box, fill=black!5, draw=black!50},  
    engine/.style={box, minimum height=0.6cm},  
    pkg/.style={draw=black!60, rounded corners=1.5pt, inner sep=3.5pt, align=center, fill=white, line width=0.7pt, minimum width=0.45cm, minimum height=0.45cm},  
    gnode/.style={circle, draw=black!50, fill=black!10, inner sep=1.5pt, line width=0.5pt},  
    unode/.style={circle, inner sep=1.8pt, line width=0.7pt},  
    gedge/.style={black!40, line width=0.6pt}, 
    arr/.style={-{Stealth[length=2.2mm,width=1.6mm]}, black!65, line width=0.8pt}, 
    ctx/.style={font=\scriptsize, text=black!75, inner sep=1pt},  
    sidetext/.style={align=center, text=black!70, font=\scriptsize},  
    panel/.style={rounded corners=6pt, fill=black!2, draw=black!15, line width=0.6pt, inner sep=6pt}
]

\node[goalbox] (ga) at (0.1,0) {goal $g$};

\node[engine, fill=FigRed!10, draw=FigRed!60!black] (enga) at (2.85,0)  
    {researcher-agnostic\\co-scientist\ \ $f_i(g,o_{<i})$};

\node[pkg] (qa3) at (5.24,0.12) {\phantom{$\mathcal{Q}$}};
\node[pkg] (qa2) at (5.17,0.06) {\phantom{$\mathcal{Q}$}};
\node[pkg] (qa1) at (5.10,0)    {$\mathcal{Q}$};

\node[sidetext] (sametext) at (6.4,0) {same for\\ \emph{every} user $u$};

\draw[arr] (ga) -- (enga);
\draw[arr] (enga) -- (qa1);

\node[anchor=west, font=\footnotesize] (laba) at (-0.8,0.7)  
    {\textbf{(a)}\, \textcolor{FigRed!80!black}{\bf \scshape Researcher-agnostic}};
    
\coordinate (ra) at (7.35,0);

\begin{scope}[yshift=-2.2cm] 
    
    \node[gnode] (a1) at (-0.30,0.30) {};
    \node[gnode] (a2) at (0.34,0.32) {};
    \node[gnode] (a3) at (-0.34,-0.26) {};
    \node[gnode] (a4) at (0.30,-0.30) {};
    \node[unode, fill=FigBlue!80, draw=FigBlue!60!black] (u1) at (0,0) {};
    
    \foreach \i in {a1,a2,a3,a4}{
        \draw[gedge] (u1)--(\i);
    }
    \draw[gedge] (a1)--(a2) (a1)--(a3) (a3)--(a4) (a2)--(a4);
    
    \node[text=FigBlue!80!black, anchor=west, font=\scriptsize] at (0.45,-0.04) {$u_1$};
    
    \node[engine, fill=FigBlue!10, draw=FigBlue!60!black] (eng1) at (2.85,0) 
        {$f_i(g,o_{<i}\mid c_{u_1})$};
        
    \node[pkg, fill=FigBlue!10, draw=FigBlue!60!black] (q1) at (5.1,0) {$\mathcal{Q}_{u_1}$};
    
    \draw[arr] (0.95,0) -- (eng1);
    \node[ctx, anchor=south east] at (1.88,0.06) {$\mathbf{z}_{u_1}\!\rightarrow\!c_{u_1}$};
    \draw[arr] (eng1) -- (q1);
    
    \node[anchor=west, font=\footnotesize] (labb) at (-0.8,0.7)  
        {\textbf{(b)}\, \textcolor{FigGreen!80!black}{\bf \scshape Personalized}};
\end{scope}

\begin{scope}[yshift=-3.8cm]
    \node[gnode] (b1) at (-0.62,0.26) {};
    \node[gnode] (b2) at (-0.62,-0.26) {};
    \node[gnode] (b3) at (-0.32,0) {};
    \draw[gedge] (b1)--(b2) (b1)--(b3) (b2)--(b3);
    
    \node[gnode] (b4) at (0.52,0.26) {};
    \node[gnode] (b5) at (0.52,-0.26) {};
    \node[gnode] (b6) at (0.20,0) {};
    \draw[gedge] (b4)--(b5) (b4)--(b6) (b5)--(b6);
    
    \node[unode, fill=FigGreen!80, draw=FigGreen!60!black] (u2) at (-0.06,0.3) {};
    \draw[gedge] (u2)--(b3) (u2)--(b6);
    
    \node[text=FigGreen!70!black, anchor=south, font=\scriptsize] at (-0.06,0.48) {$u_2$};
    
    \node[engine, fill=FigGreen!10, draw=FigGreen!60!black] (eng2) at (2.85,0) 
        {$f_i(g,o_{<i}\mid c_{u_2})$};
        
    \node[pkg, fill=FigGreen!10, draw=FigGreen!60!black] (q2) at (5.1,0) {$\mathcal{Q}_{u_2}$};
    
    \draw[arr] (0.95,0) -- (eng2);
    \node[ctx, anchor=south east] at (1.88,0.06) {$\mathbf{z}_{u_2}\!\rightarrow\!c_{u_2}$};
    \draw[arr] (eng2) -- (q2);
\end{scope}

\node[goalbox] (gb) at (2.85,-2.95) {same goal $g$};
\draw[arr] (gb) -- (eng1);
\draw[arr] (gb) -- (eng2);

\draw[densely dashed, black!50, line width=0.8pt] (q1.south) -- (q2.north);
\node[fill=black!2, inner sep=2pt, text=black!85, font=\footnotesize] at (5.1,-2.95) {$\neq$};

\node[sidetext] (difftext) at (6.4,-2.95) {different for\\ \emph{each} user $u$};
\coordinate (rb) at (7.35,-2.95);

\begin{scope}[on background layer]
    \node[panel, fit=(laba)(ga)(enga)(qa3)(sametext)(ra)] {};
    \node[panel, fit=(labb)(u1)(a3)(eng1)(q1)(b2)(eng2)(q2)(difftext)(gb)(rb)] {};
\end{scope}

\end{tikzpicture}
\caption{%
\textbf{(a)}~A researcher-agnostic co-scientist maps the same abstract goal $g$ to an identical research package $\mathcal{Q}$ for every researcher. \textbf{(b)}~Personalized auto-research conditions every stage on a graph-derived context $c_u$. For example, researcher $u_1$ lies in a dense network cluster, while $u_2$ bridges a structural hole. Consequently, the same goal naturally yields distinct evidence, search trajectories, and output packages ($\mathcal{Q}_{u_1} \neq \mathcal{Q}_{u_2}$). Each package is thereby optimized for feasibility, alignment, and novelty relative to the individual researcher's 
domain.}
\label{fig:teaser}
\vspace{-4mm}
\end{figure}

\section{Introduction}
\label{sec:introduction}
Scientific publication has grown exponentially for decades \citep{wang2021science}, and no individual researcher can fully navigate the literature of their own field, let alone integrate insights from adjacent disciplines. This fundamental problem motivates the recent line of work on AI co-scientists and auto-research systems, that is, language-model agents that generate hypotheses, retrieve and synthesize related work, design and execute experiments, and draft manuscripts~\citep{lu2024aiscientist,gottweis2025coscientist,yamada2025aiscientistv2,schmidgall2025agentlab,tang2025airesearcher}. Since the 2024 AI Scientist prototype, the field has moved rapidly to progressive agentic tree search, experiment-manager control, shared preprint-style archives, cross-run memory, human-in-the-loop co-research, and explicit provenance and safety mechanisms \citep{schmidgall2025agentrxiv,zhang2025aixiv,tie2026autoresearchai,liu2026autoresearchclaw,wang2026parness,jeddi2026gear}. 
 
Despite their importance and rapid progress, these systems share a fundamental structural limitation: they are \emph{researcher-agnostic}. Given the same goal, they produce the same distribution of outputs whether the requester is a first-year doctoral student or a senior professor, a graph-mining researcher or a computational biologist (Figure~\ref{fig:teaser}). The output may be competent, but it is also interchangeable, and this interchangeability contradicts how scientific ideas are often discovered. Intuitively, novel directions frequently arise from a scientist's idiosyncratic combination of prior failures, methodological taste, and hard-won experimental intuition, and a homogeneous system erases precisely the heterogeneity that makes research creative. 
The stakes span every level: (i) \emph{epistemic}, since most scientific capability is tacit and absent from the literature, and is thus unreachable by any literature-conditioned system; (ii) \emph{systemic}, since many researchers querying one generic engine drives the field toward a scientific monoculture, racing the same ideas while counterfactually valuable directions go unexplored; and (iii) \emph{practical}, since researchers can only verify, and will only adopt, directions matched to their expertise and resources.

In this work, we introduce \emph{personalized auto-research}, the problem of conditioning every stage of the research process on a representation of the individual researcher. Notably, this problem is fundamentally different from both existing AI co-scientist systems and prior work on personalized language models \citep{salemi2024lamp,ouyang2022instructgpt}: existing co-scientists automate research but largely ignore the researcher, whereas existing personalization work models user-specific outputs but not the sequence of scientific decisions that constitute research. In personalized auto-research, the object being personalized is an end-to-end research trajectory, not merely a response.
 
\smallskip\noindent\textbf{Summary of contributions.} The key contributions of this work are as follows:
\textbf{(1)}~We formalize the problem of personalized auto-research 
(\S\ref{sec:problem}).
\textbf{(2)}~We propose a general and flexible framework (Algorithm~\ref{alg:personalized-autoresearch}) for this problem that personalizes every step such as
retrieval, hypothesis search, experimentation, writing, citation, and review, etc
(\S\ref{sec:method}).
\textbf{(3)}~We present a vision for personalized auto-research organized around three fundamental components: researcher representation, personalization across the full research pipeline, and evaluation grounded in the individual (\S\ref{sec:vision}).
\textbf{(4)}~Finally, we discuss open problems and challenges.
(\S\ref{sec:challenges}).

\section{Background}\label{sec:background}
\noindent\textbf{AI Co-Scientists and Auto-Research.}
Long-horizon reasoning and tool use \citep{wei2022cot,yao2023react} have enabled agents that carry out research end-to-end. The AI Scientist established the autonomous loop of proposing, implementing, writing up, and reviewing experiments \citep{lu2024aiscientist,lu2026endtoend}, and AI Scientist-v2 replaced its template-dependent loop with progressive agentic tree search, an experiment manager, parallel execution, and vision-language figure feedback \citep{yamada2025aiscientistv2}. A growing set of systems extends this across multi-agent hypothesis generation, staged human-feedback workflows, automated data science, and long-horizon discovery \citep{gottweis2025coscientist,gottweis2026coscientistnature,schmidgall2025agentlab,tang2025airesearcher,yang2025rdagent,weng2025deepscientist,mitchener2025kosmos,pu2025piflow}. A parallel line builds the surrounding infrastructure, namely shared archives, cross-run memory, and population-level search over code states \citep{schmidgall2025agentrxiv,zhang2025aixiv,wang2026parness,liu2026autoresearchclaw,jeddi2026gear,tie2026autoresearchai}, and a third studies evaluation and risk through discovery benchmarks, manuscript verification, safety, and critiques of implementation and evaluation bottlenecks \citep{chen2025autobench,luo2025baisbench,son2025spota,zhu2025safescientist,zhu2025implementation,luo2025hiddenpitfalls}; recent surveys organize the space by research stage and autonomy level \citep{luo2025llm4sr,eger2025transforming,zheng2025automation,ren2025scientificintelligence,wei2025agenticscience,zhou2025autonomousagents}.

Auto-research is thus no longer speculative, spanning agentic search, automated implementation, full-paper generation, and safety controls, with domain systems such as AlphaFold showing how AI complements expert judgment in high-stakes science \citep{jumper2021alphafold}. Yet the dominant objective is task-, benchmark-, or field-conditioned: quality is optimized with respect to the goal, literature, evidence, or reviewer model, never with respect to the individual researcher who will adopt the output. Even scientist-in-the-loop systems \citep{gottweis2025coscientist,schmidgall2025agentlab} condition only on manual, session-level guidance, so interaction is not personalization, which requires a learned, persistent representation of what the scientist cannot articulate in a prompt. Auto-research and co-scientist are therefore not synonyms: the former names a capability (the automation of research) while the latter names a relationship with a specific researcher. Table~\ref{tab:landscape} makes this explicit: personalization is orthogonal to the autonomy axis along which the field has advanced, and the bottom-right quadrant, where interaction updates the personalization itself, is the true AI co-scientist we pose and the gap this work addresses.

\begin{table}[t]
\caption{Personalization is orthogonal to autonomy. Existing systems advance along the autonomy axis while remaining researcher-agnostic, whereas our vision of \textcolor{googleblue}{\bf Personalized Auto-Research} (Alg.~\ref{alg:personalized-autoresearch}) and our \textbf{\textcolor{googlegreen}{True Personalized AI Co-Scientist}} (Alg.~\ref{alg:personalized-autoresearch} with human-in-the-loop).
}
\label{tab:landscape}
\vspace{-2mm}
\footnotesize
\setlength{\tabcolsep}{5pt}
\renewcommand{\arraystretch}{1.99}
\begin{tabular}{
>{\raggedright\arraybackslash}p{0.165\columnwidth}
>{\raggedright\arraybackslash}p{0.34\columnwidth}
>{\raggedright\arraybackslash}p{0.365\columnwidth}
}
\toprule
& \textbf{\scshape Researcher-agnostic} & \textbf{\scshape Personalized} \\
\midrule
\textbf{Fully autonomous}
& AI Scientist~\citep{lu2024aiscientist,yamada2025aiscientistv2}, DeepScientist~\citep{weng2025deepscientist}, ...
& 
\textbf{\textcolor{googleblue}{Personalized Auto-Research 
}} (Alg.~\ref{alg:personalized-autoresearch})
\\
\textbf{Human-in-the-loop}
& Co-Scientist \citep{gottweis2025coscientist,gottweis2026coscientistnature}, 
AutoResearchClaw~\citep{liu2026autoresearchclaw}, ...
& 
\textbf{\textcolor{googlegreen}{True AI Co-Scientist}} 
(Alg.~\ref{alg:personalized-autoresearch} w/ human-in-the-loop)
\\
\bottomrule
\end{tabular}
\vspace{-2mm}
\end{table}

\medskip\noindent\textbf{Personalization.}
Recommender systems learn latent user representations from interactions \citep{koren2009matrix}, while preference alignment \citep{ouyang2022instructgpt}, retrieval-augmented generation \citep{lewis2020rag}, and the LongLaMP benchmark \citep{kumar2024longlamp} establish that user-conditioned language modeling is tractable and beneficial. 
However, personalizing a recommendation or a single output is far narrower than personalizing science. In personalized auto-research, the object is a sequence of decisions: what to retrieve, what to hypothesize, which experiments to run, how to frame the write-up, 
how to revise the resulting artifact, etc.
 
 
\section{Problem Formulation}
\label{sec:problem}
More formally, let $\mathcal{U}$ denote the population of researchers and let
$\mathcal{G} = (\mathcal{V}, \mathcal{E}_{\mathcal{G}}, \tau_V, \tau_E)$
denote a heterogeneous graph over the research landscape, where $\mathcal{V}$ contains researcher, paper, venue, institution, method, dataset, and topic nodes; $\mathcal{E}_{\mathcal{G}}$ contains co-authorship, citation, publication, affiliation, usage, and topic-assignment edges; and $\tau_V,\tau_E$ assign node and edge types ($\mathcal{E}_{\mathcal{G}}$ is distinct from the literature index $\mathcal{I}$ used below). For a researcher $u \in \mathcal{U}$, let $\mathcal{S}_u$ denote observed signals (papers, citations, code, review history, venue preferences, explicit constraints), let $\mathbf{z}_u \in \mathbb{R}^d$ be a graph-derived representation of $u$, and let $c_u$ be the operational context given to the auto-research system:
\begin{equation}
\mathbf{z}_u = \mathrm{Enc}_{\mathcal{G}}(u; \mathcal{G}), \qquad
c_u = \Phi(\mathcal{S}_u,\mathbf{z}_u).
\label{eq:context}
\end{equation}
 
\begin{definition}[\textbf{\scshape Personalized Auto-Research}]
\label{def:par}
\it
Let $g$ be a research goal and $\mathcal{P}=(p_1,\ldots,p_K)$ the stages of the research process (literature retrieval, hypothesis generation, experiment design, code execution, writing, citation, refinement, review). Whereas a researcher-agnostic co-scientist parameterizes each stage by the goal alone, $o_i=f_i(g,o_{<i})$, personalized auto-research learns stage models conditioned on the researcher,
\begin{equation}
o_i = f_i(g, o_{<i} \mid c_u),
\label{eq:stage}
\end{equation}
where $o_{<i}$ denotes outputs of earlier stages.
\end{definition}
 
The desiderata are threefold: the output should be (i) \emph{feasible}, respecting the researcher's capabilities, resources, and constraints; (ii) \emph{aligned}, compatible with the researcher's scientific identity, community, and style; and (iii) \emph{novel}, new to the field and distinct from the researcher's prior work. Feasibility and alignment are personalized properties, whereas novelty is partly field-level and partly user-relative. This distinction is fundamentally important, since without it, personalization collapses into a recommender that predicts more of the same.
 
\section{Personalized Auto-Research}
\label{sec:method}
Algorithm~\ref{alg:personalized-autoresearch} gives the proposed personalized auto-research procedure. The key idea is to formulate personalization as a modification of the SOTA agentic auto-research loop, that is, agentic tree search over partial research states with an experiment manager, code execution, figure refinement, manuscript writing, automated review, and provenance logging \citep{yamada2025aiscientistv2,liu2026autoresearchclaw,wang2026parness,jeddi2026gear}, rather than the linear 2024 template loop. Personalization adds a user $u$, a context $c_u$, a personalized evidence set $\mathcal{R}_u(g)$, and a user-conditioned utility, where retrieval scores documents $d$ in corpus $\mathcal{D}$ jointly by goal and researcher and $\TopOp_k$ returns the $k$ highest-scoring items:
\begin{align}
s_u(d \mid g) &= \big\langle \eta(d),\, \rho(g,c_u) \big\rangle,
\quad
\mathcal{R}_u(g) = \TopOp_k\!\big(\mathcal{D}; s_u(\cdot\mid g)\big),
\label{eq:retrieve} \\
U(h \mid g,u) &=
\alpha\Nov(h,\mathcal{I},c_u)
+\beta\Rel(h,g,c_u)
+\gamma\Feas(h,\mathcal{W}_0,c_u).
\label{eq:utility}
\end{align}
Notably, Eq.~\eqref{eq:utility} is one key 
distinction: hypotheses are \emph{ranked} by 
personalized 
novelty, relevance, and feasibility, rather than filtered by a binary, global novelty test.
Intuitively, the same idea can be infeasible for one researcher, obvious to another, and transformative for a third whose graph position makes a new bridge credible.
 
\definecolor{StageBlue}{HTML}{4285F4}
\definecolor{StageGreen}{HTML}{34A853}
\definecolor{StageRed}{HTML}{EA4335}
\newcommand{\AlgSection}[2]{%
\vspace{2mm}
    \Statex \textcolor{#1}{\textbf{// #2}}
}
\begin{algorithm}[t]
\caption{\;Personalized Auto-Research}
\label{alg:personalized-autoresearch}
\scriptsize
\begin{algorithmic}[1]
\Require language-model agents $\pi$, experiment manager $\mu$, vision-language reviewer $\omega$, research goal $g$, initial workspace $\mathcal{W}_0$, seed archive $\mathcal{J}$, literature index $\mathcal{I}$ over corpus $\mathcal{D}$, research graph $\mathcal{G}$, user $u$ with signals $\mathcal{S}_u$, context encoder $\Phi$, tree-search budget $N$, branch factor $b$, package count $m$, exp. budget $B$
\Ensure 
personalized reprod.\ research packages $\mathcal{A}_u$
\AlgSection{StageBlue}{Personalized Context and Literature Grounding}
\State $\mathbf{z}_u \gets \mathrm{Enc}_{\mathcal{G}}(u;\mathcal{G})$ \label{line:embed}
\State $c_u \gets \Phi(\mathcal{S}_u,\mathbf{z}_u)$ \label{line:context}
\State $\mathcal{R}_u(g) \gets \TopOp_k(\mathcal{D};s_u(\cdot\mid g))$ \label{line:retrieve}
\State $\mathcal{B}_u \gets \{(\mathcal{W}_0,\emptyset,\emptyset,\emptyset,0)\}$ \label{line:initpool}
\AlgSection{StageGreen}{Personalized Hypothesis Search and Experimentation}
\For{$r = 1$ \textbf{to} $N$} \label{line:treeloop}
    \State $x \gets \mu(\texttt{select}, \mathcal{B}_u, g,\mathcal{R}_u(g),c_u)$ \label{line:selectstate}
    \State $\mathcal{X} \gets \pi(\cdot \mid \texttt{expand}, x,g,\mathcal{J},\mathcal{R}_u(g),c_u,b)$ \label{line:expand}
    \For{each child state $x' \in \mathcal{X}$} \label{line:childloop}
        \State $h \sim \pi(\cdot \mid \texttt{hypothesize}, x',g,\mathcal{J},\mathcal{R}_u(g),c_u)$ \label{line:propose}
        \State $U(h\mid g,u) \gets \alpha\Nov(h,\mathcal{I},c_u)+\beta\Rel(h,g,c_u)+\gamma\Feas(h,\mathcal{W}_0,c_u)$ \label{line:utility}
        \State $\mathcal{C} \sim \pi(\cdot \mid \texttt{implement}, x',h,\mathcal{W}_0,c_u)$ \label{line:edit}
        \State $(\mathcal{O},\mathcal{F},\Lambda_u) \gets \Execute(\mathcal{C},h,B,c_u)$ \label{line:execute}
        \State $\mathcal{F} \sim \omega(\cdot \mid \texttt{figure-review}, \mathcal{F},\mathcal{O},g,c_u)$ \label{line:figure}
        \State $q_u \gets \mu(\texttt{score}, U(h\mid g,u),h,\mathcal{C},\mathcal{O},\mathcal{F},g,c_u)$ \label{line:score}
        \State $\mathcal{B}_u \gets \mathcal{B}_u \cup \{(h,\mathcal{C},\mathcal{O},\mathcal{F},q_u,\Lambda_u)\}$ \label{line:addpool}
    \EndFor
    \State \textbf{optionally:} $e_r \gets u\big(\texttt{feedback}, \TopOp_1(\mathcal{B}_u;q_u)\big)$,\; $\mathcal{S}_u \gets \mathcal{S}_u \cup \{e_r\}$,\; $c_u \gets \Phi(\mathcal{S}_u,\mathbf{z}_u)$ \label{line:feedback}
\EndFor
\AlgSection{StageRed}{Personalized Research Package Synthesis}
\State $\mathcal{A}_u \gets \emptyset$ \label{line:initarchive}
\For{each state $(h,\mathcal{C},\mathcal{O},\mathcal{F},q_u,\Lambda_u) \in \TopOp_m(\mathcal{B}_u;q_u)$} \label{line:paperloop}
    \State $y \sim \pi(\cdot \mid \texttt{write}, g,h,\mathcal{C},\mathcal{O},\mathcal{F},\mathcal{R}_u(g),c_u)$ \label{line:write}
    \State $y \sim \pi(\cdot \mid \texttt{cite-refine}, y,\mathcal{R}_u(g),\mathcal{O},\mathcal{F},c_u)$ \label{line:refine}
    \State $v \sim \pi(\cdot \mid \texttt{review}, y,g,\mathcal{O},\mathcal{F},c_u)$ \label{line:review}
    \State $\mathcal{Q}_u \gets (h,U(h\mid g,u),\mathcal{C},\mathcal{O},\mathcal{F},y,v,\Lambda_u,c_u,\mathcal{R}_u(g))$ \label{line:package}
    \State $\mathcal{A}_u \gets \mathcal{A}_u \cup \{\mathcal{Q}_u\}$ \label{line:archive}
\EndFor
\State \Return $\mathcal{A}_u$ \label{line:return}
\end{algorithmic}
\end{algorithm}
 
\paragraph{Personalized context and evidence}
Lines~\ref{line:embed}--\ref{line:retrieve} construct the user-specific state: the graph encoder produces $\mathbf{z}_u$, the context encoder produces $c_u$, and retrieval produces $\mathcal{R}_u(g)$. Notably, this evidence set is not simply the topically closest literature to $g$. It is the one that is relevant to the goal \emph{and} useful for this researcher, given their prior work, collaborators, methods, resources, and position in $\mathcal{G}$.
 
\paragraph{Personalized hypothesis search}
Lines~\ref{line:treeloop}--\ref{line:addpool} replace a universal agentic tree with a user-conditioned tree $\mathcal{B}_u$: the manager selects partial states, the language model expands them, and every hypothesis is scored by $U(h\mid g,u)$, all under $c_u$. Two researchers with the same goal therefore induce \emph{different search trees}, since their contexts change the selection policy, the expansion distribution, and the utility. Furthermore, candidates are implemented and executed under the user's constraints (Lines~\ref{line:edit}--\ref{line:execute}), and even vision-language figure feedback may depend on $c_u$ (Line~\ref{line:figure}): a graph-mining researcher and a biomedical collaborator need different visual encodings and terminology for the same quantitative result. The researcher also remains in the loop: at any iteration, $u$ may optionally critique the current best state (Line~\ref{line:feedback}). Notably, this feedback does more than steer the session, as in existing human-in-the-loop systems \citep{gottweis2025coscientist,schmidgall2025agentlab,liu2026autoresearchclaw}, where guidance evaporates when the session ends. Here it is folded into $\mathcal{S}_u$ and re-encoded into $c_u$, and thus the representation of the relationship itself is updated, both within a run and across runs.
 
\paragraph{Personalized package synthesis}
Lines~\ref{line:paperloop}--\ref{line:return} write, cite, refine, and review the top-$m$ states under the same context. Each terminal artifact is a reproducible research package
$\mathcal{Q}_u=(h,U(h\mid g,u),\mathcal{C},\mathcal{O},\mathcal{F},y,v,\Lambda_u,c_u,\mathcal{R}_u(g))$
consisting of the hypothesis, its personalized score, code, outputs, figures, paper, automated review, provenance log, and the context and evidence that conditioned the run, that is, the metadata needed to audit \emph{why this package was produced for this researcher}. 
 
 
\section{Personalized Auto-Research Vision}
\label{sec:vision}
The distinction between an instrument and a collaborator is not one of capability but of relationship: a capable instrument returns high-quality outputs, while a collaborator returns outputs tailored to the person it works with. Personalization is therefore not a marginal improvement to AI co-scientists, but rather what makes the metaphor accurate. We organize the resulting research agenda around three fundamental components.
 
\paragraph{Researcher Representation.}\label{sec:pillar-representation}
A researcher's publications alone are a thin description of their scientific identity. Intuitively, two researchers with similar publication lists can have fundamentally different research identities if one lies in a dense cluster of theorists while the other bridges that cluster to computational biology. This leads us to propose learning $\mathbf{z}_u$ from the researcher's position in $\mathcal{G}$, aggregating multi-hop neighborhoods that capture who their collaborators work with, where their extended network publishes, and which topics are adjacent to their community. Notably, this grounds personalization in structural properties of science: the most valuable direction for a researcher is often not an extension of their work but a \emph{structural hole}, that is, a bridge between a region they inhabit and a nearby region not yet connected \citep{burt2004structural}.
 
\paragraph{Pipeline Personalization.}
\label{sec:pillar-process}
Personalizing only hypothesis generation while leaving retrieval, experiment design, writing, citation, and review researcher-agnostic is internally inconsistent: a hypothesis can be aligned yet require resources the researcher lacks and cite a literature that omits their community. The design principle is to separate \emph{exploitation}, which extends the researcher's trajectory, from \emph{exploration}, which uses the profile to judge feasibility and framing while keeping the candidate space broad. The goal is to personalize the \emph{how} without narrowing the \emph{what}. Personalization also need not stop at the stages: the components Algorithm~\ref{alg:personalized-autoresearch} treats as fixed can themselves be learned per researcher. The weights in Eq.~\eqref{eq:utility} and the $\mathrm{Nov}$, $\mathrm{Rel}$, and $\mathrm{Feas}$ functions can be fit from research traces such as revisions, reviews, and abandoned versus published projects; feasibility can be grounded in the researcher's actual repositories, compute, and datasets rather than a self-reported profile, and review calibrated to what they can rigorously verify, keeping autonomous science inside their expertise and auditable.
 
\paragraph{Evaluation Grounded in the Individual.}
\label{sec:pillar-evaluation}
Predicting a researcher's next papers is a flawed proxy: held-out papers record what they happened to work on under path-dependent incentives, not what they \emph{should} have, and the proxy penalizes a co-scientist that recommends something better than anything they pursued. Evaluation should instead combine (i) \emph{feasibility alignment}, whether directions are executable given documented capabilities; (ii) \emph{expert-assessed quality}, whether blinded experts judge ideas novel, significant, and appropriate for the profile; and (iii) \emph{longitudinal impact}, whether co-scientist use yields more influential work over multi-year horizons. Held-out papers nonetheless serve as a scalable \emph{necessary condition}: holding out a paper at time $t$ and building $c_u$ from the record prior to $t$, one conditions the system on the general concept and measures \emph{fidelity}, whether it recovers a hypothesis and experimental path close to what the researcher pursued, and \emph{contrast}, whether the researcher-agnostic system stays generic while different contexts diverge on the same concept. High fidelity with high contrast shows $c_u$ carries signal; failure falsifies the representation. Fully automatable from bibliographic data, this protocol validates the personalization signal rather than the quality ceiling, and building benchmark infrastructure for all of these criteria is itself a major missing contribution.

\section{Open Challenges}
\label{sec:challenges}
We now discuss key open problems and challenges, which are deep tensions in the problem itself rather than engineering obstacles.
 
\paragraph{Creativity Collapse.}\label{sec:creativity-collapse}
A researcher-agnostic system applies one map from goal to output distribution, so the epistemic loss compounds at population scale: when many researchers query the same engine, the field's portfolio of explored hypotheses collapses toward a monoculture, and globally ``best'' ideas are raced redundantly while directions of high counterfactual value, those only a particular researcher is positioned to pursue, go unexplored. Personalization is therefore a decorrelation mechanism for collective discovery, not merely a convenience for individuals. The challenge is to exploit a researcher's experience without collapsing into biographical mimicry, treating it as a signal for experiments a generic system would not propose yet this researcher can develop. This calls for rewarding \emph{counterfactual complementarity}: ideas unlikely under both a universal model and the user's past work alone, yet plausible and valuable given their accumulated experience. One realization trains a mimicry model of the researcher and optimizes for high utility $U(h\mid g,u)$ at low likelihood under both the mimicry and universal models.

\paragraph{Lifecycle Dependence and Cold Start.}
\label{sec:lifecycle}
For a senior researcher with a dense graph, value lies in adjacent unexplored territory and bridging structural holes, whereas for an early-career researcher with sparse history, the system must help \emph{establish} an identity rather than extend one. Notably, this exceeds the classical cold-start problem, since the objective function itself changes with career stage.
 
\paragraph{Team-Personalized Auto-Research.}
\label{sec:team}
Most impactful research is carried out by teams rather than individuals \citep{wuchty2007teams}, yet the formulation above personalizes for a single researcher. The framework extends naturally by replacing $u$ with a team $T \subseteq \mathcal{U}$ and defining $c_T = \Psi(\{(\mathcal{S}_u,\mathbf{z}_u)\}_{u \in T})$, where a team is simply a subgraph of $\mathcal{G}$. Notably, the desiderata aggregate asymmetrically: feasibility is a union, since the team can execute what any member can execute; novelty is relative to the union of prior work; and alignment is closer to an intersection, since framing must be evaluable by every community the team spans. This asymmetry is precisely why scientists collaborate, and it enables fundamentally new capabilities, including routing experiments to the member best positioned to execute them, writing and citing for the union of the team's communities, and recommending the collaborator whose addition closes the structural hole that makes a hypothesis credible \citep{burt2004structural}. Notably, Algorithm~\ref{alg:personalized-autoresearch} supports this setting with only local modifications: Lines~\ref{line:embed}--\ref{line:context} encode each member and aggregate via $\Psi$ to obtain $c_T$; retrieval (Line~\ref{line:retrieve}) scores documents against the team context; the utility (Line~\ref{line:utility}) becomes $U(h \mid g,T)$ with $\Feas(h,\mathcal{W}_0,c_T)=\max_{u\in T}\Feas(h,\mathcal{W}_0,c_u)$ and alignment taken as a minimum over members; implementation and execution (Lines~\ref{line:edit}--\ref{line:execute}) assign each candidate to $\arg\max_{u\in T}\Feas(h,\mathcal{W}_0,c_u)$; and the feedback step (Line~\ref{line:feedback}) collects critiques from multiple members, updating each $\mathcal{S}_u$ and re-aggregating $c_T$. However, aggregating conflicting member preferences into a single $c_T$ is a social-choice problem, the max-based feasibility assumes frictionless handoffs between members, and multi-party privacy becomes harder when the user is itself a group.
 
\paragraph{Evaluation Without Ground Truth.}
\label{sec:no-ground-truth}
The fundamental difficulty is that the quantity we need to evaluate, namely the value of a recommended direction for a specific researcher, is never observed. History records only the single trajectory each researcher actually followed, and that trajectory was shaped by funding, advisors, reviewing, and chance rather than by an oracle over alternatives. Scoring a system by similarity to this record therefore rewards mimicry and penalizes better recommendations (\S\ref{sec:pillar-evaluation}). The held-out protocol of \S\ref{sec:pillar-evaluation} confirms only a necessary condition, namely that $c_u$ carries researcher-specific signal; it cannot certify that a recommended direction is \emph{good}, since the value of an unpursued alternative is never recorded. Resolving that distinction requires human judgment (expert panels), time (longitudinal studies), or interventions (counterfactual designs), each of which necessitates further research.

\section{Conclusion}\label{sec:conclusion}
We introduced \emph{personalized auto-research}, the problem of conditioning the full research process on a representation of the individual researcher. The core claim is simple: a system cannot be a true co-scientist if it does not know whom it is collaborating with. The goal is complementarity rather than similarity, just as scientists choose collaborators for the expertise they lack, and so a stronger universal engine does not resolve the problem; it still returns the same high-scoring package to everyone, which makes personalization orthogonal to, and required on top of, the current SOTA. We therefore pose personalized auto-research as a grand challenge for the AI ecosystem, spanning graph learning for representation, agentic systems for execution, HCI for consent and control, and community benchmarks for individual-grounded evaluation. No single group can deliver it alone, and the tensions it raises around novelty, equity, privacy, lifecycle, and evaluation define the field as much as its algorithms do.
 
\bibliographystyle{ACM-Reference-Format}
\bibliography{paper}
\end{document}